\documentclass[11pt]{article}

\usepackage[final]{acl}

\usepackage{times}
\usepackage{latexsym}

\usepackage[T1]{fontenc}
\usepackage[utf8]{inputenc}

\usepackage{microtype}

\usepackage{inconsolata}

\usepackage{graphicx}

\usepackage{graphicx}
\usepackage{enumitem}

\newcommand{\GP}[1]{$\mathcal{P}_{#1}$}

\usepackage{CJKutf8}

\usepackage{amsmath}  % 数学公式
\usepackage{booktabs} % 专业表格
\usepackage{graphicx} % 图片支持
\usepackage{amsmath,amssymb} % 必需数学宏包
\usepackage{float} % 在导言区添加 表格置顶
\title{Poller: Are LLMs Suitable for Evaluating the Poetry Understanding Task?}

\author{
       Shanshan Wang$^1$~~~~
        Derek F. Wong$^1\thanks{Corresponding Author}$~~~~
        Jingming Yao$^2$~~~~
        \textbf{Lidia S. Chao$^1$}~~~~
       \\
    $^1$NLP$^2$CT Lab, Department of Computer and Information Science, University of Macau \\
    {nlp2ct.shanshan@gmail.com, \{derekfw,lidiasc\}@um.edu.mo} \\
     $^2$Department of Portuguese, Faculty of Arts and Humanities, University of Macau\\
       {jmyao@um.edu.mo} \\
    }

\begin{document}
\maketitle
\begin{abstract}
% Traditional automatic evaluation methods have been shown to be unsuitable for modern Chinese poetry due to the unique characteristics of poetry itself. 

Traditional automatic evaluation methods have been shown to be unsuitable for modern Chinese poetry because of the distinct nature of this literary genre. Human evaluation remains reliable, but is expensive and not applicable to large-scale data. In this paper, we propose Poller (Poetry LLM Evaluator), a novel method leveraging large language models (LLMs) to evaluate the poetry understanding task. Specifically, our method requires LLMs to play the role of a poem's author with detailed information, thereby emulating human evaluation and judgment by adopting the poet's perspective. We conducted comprehensive experiments on multiple LLMs, evaluating the interpretations of poems across eight specialized dimensions. Experimental results demonstrate that our method effectively reduces the evaluation error between LLMs and humans. Especially for specific dimension evaluation, Poller-based LLMs achieve a 94.55\% and 89.53\% error reduction for rhetorical techniques and defamiliarization, respectively, compared to baseline methods. These performances are unattainable by conventional LLM evaluation methods. Experimental results from multiple LLMs across various dimensions validate the efficacy of our method. This work bridges the gap between automated efficiency and human expertise, establishing a foundation for automated evaluation in poetry-related tasks.

% Traditional automatic evaluation methods have been shown to be unsuitable for modern Chinese poetry due to the unique characteristics of poetry itself. Human evaluation remains reliable, but it is expensive and not applicable to large-scale data. In this paper, we propose Poller (Poetry LLM Evaluator), a novel method leveraging Large Language Models (LLMs) to evaluate poetry understanding task. Specifically, our method requires LLMs to play the role of poem's author with detailed information (e.g., basic information, poetic views, and critics' comments), thereby emulating human evaluation and judgment by adopting the poet's perspective. We conducted comprehensive experiments on multiple LLMs, evaluating the interpretations of poems across eight specialized dimensions. Experimental results demonstrate that our Poller method effectively reduces the evaluation error between LLMs and humans. Especially for specific dimension evaluation, Poller-based LLMs achieve a 94.55\% reduction in error for rhetorical techniques and an 89.53\% reduction in error for defamiliarization compared to baseline methods. These performances are unattainable by conventional LLM evaluation methods. Experimental results of multiple LLMs from multiple dimensions validate the efficacy of our method. This work bridges the gap between automated efficiency and human expertise, establishing a foundation for automated evaluation in poetry-related tasks.

\end{abstract}

\section{Introduction}

% Poetry is one of the most important literary forms. It has influenced the development of different countries, nationalities, and cultures, and plays an important role in real life \cite{simecek2016uses, yi2018chinese}. In recent years, poetry has attracted the attention of researchers in the natural language processing (NLP), mainly focusing on poetry generation and translation \cite{genzel2010poetic, greene2010automatic, ghazvininejad2018neural, antar2023effectiveness, refaee2023okaz, wang2024best, deng2024can}.

% Poetry has attracted the attention of researchers in natural language processing (NLP), mainly focusing on poetry generation and translation \cite{genzel2010poetic, ghazvininejad2018neural, refaee2023okaz, wang2024best, deng2024can}.

Poetry has attracted the attention of researchers in natural language processing (NLP), with topics such as poetry understanding \cite{wang-etal-2026-chatgpt}, generation \cite{hutson2023poetry,antar2023effectiveness,deng2024can}, detection \cite{wang-etal-2025-benchmarking,li2026wrote}, and translation \cite{virvou2023chatgpt, song2023towards, wang2024best}.

\begin{figure}[t]
    \centering
    \includegraphics[width=0.475\textwidth]{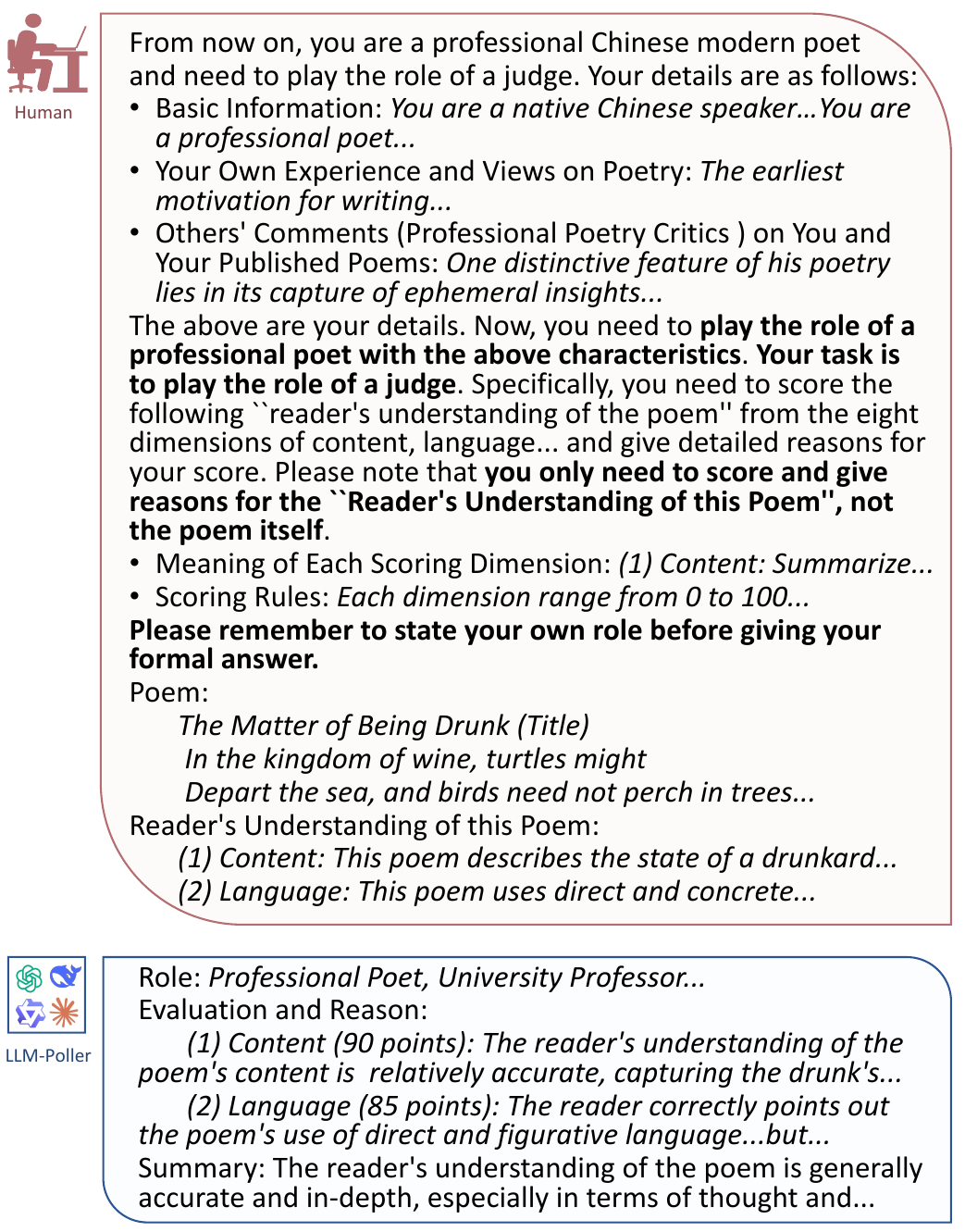}
    \caption{An example of our proposed Poller (Poetry LLM Evaluator) method.}
    \label{Example picture}
\end{figure}

\begin{figure*}[t]
    \centering
    \includegraphics[width=\textwidth]{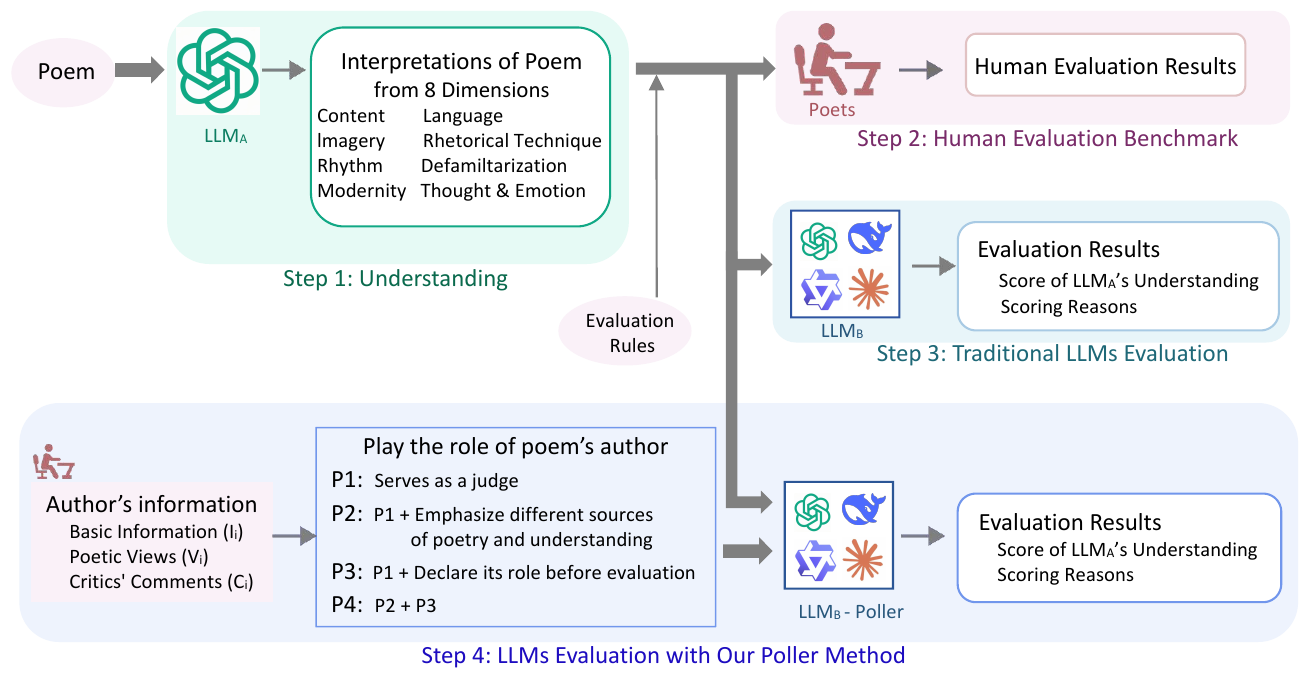}
    \caption{The overall workflow of our work, consisting of four components. Poller is the core method we propose.}
    \label{overall workflow}
\end{figure*}

% Effective and reliable evaluation is indispensable for promoting the development and progress of NLP \cite{papineni2002bleu, sun2019fine, godbole2023deep, wang2023pandalm}.

Effective and reliable evaluation is indispensable for promoting the development and progress of NLP \cite{papineni2002bleu, sun2019fine, wang2023pandalm}. Different tasks and domains require distinct evaluation criteria \cite{li2024llms}. While automatic evaluation is prevalent in NLP, it is not well-suited for poetry-related tasks due to the unique characteristics of poetry, such as its use of figurative language, rhythm, and imagery \cite{refaee2023okaz, wang2024best, novikova2017we}. In addition, the reliability of static evaluation methods may decrease when the distribution shifts occur \cite{shen2026preconditioned}. Human evaluation, often regarded as the gold standard, serves as the reference for automatic evaluation methods \cite{resnik2010evaluation}, particularly for creative tasks like poetry. Previous poetry-related works relied heavily on human evaluation \cite{manurung2012using, zhang2014chinese, yan2016, yi2018automatic}. Although human evaluation is most suitable for poetry-related tasks, it is costly and not applicable to large-scale data \cite{papineni2002bleu, liu2024llms}. The shortcomings of existing automatic and human evaluation methods have hindered the progress of poetry-related tasks. Therefore, it is urgent to propose new evaluation methods.

Fortunately, the emergence of large language models (LLMs) makes new evaluations possible, although LLMs have limitations on certain tasks \cite{lin-etal-2025-large}. LLMs were used as evaluation tools due to their obvious potential and advantages \cite{chiang2023can, wang2023chatgpt, liu2023g}, and have made effective progress in the evaluation of multiple tasks such as open-domain conversations \cite{lin2023llm}, text summarization \cite{chu2024pre}, dialogue response generation, and open-ended question answering \cite{chan2023chateval}. However, evaluation based on LLMs has the potential for bias \cite{liu2024llms, bitton2023visit, koo2023benchmarking, lan-etal-2025-f2bench, lan2025mcbe}. For example, previous text summarization task \cite{panickssery2024llm} found that LLMs, including GPT-4 and Llama 2, tend to give higher scores to their own generation.

%Previous studies have demonstrated the effectiveness of LLMs in evaluating NLP tasks. 

% For example, \citet{panickssery2024llm} used LLMs to evaluate the text summarization task and found that LLMs including GPT-4 and Llama 2 prefer and give higher scores to their own generation.

% In recent works, LLMs have been used as evaluation tools for various tasks due to their obvious potential and advantages \cite{chiang2023can, wang2023chatgpt, liu2023g}, such as the ability to produce interpretable evaluations \cite{li2024llms}.

% \citet{liu2023g} evaluated text summarization and dialogue generation using GPT-4, arguing that LLM evaluation of LLM-generated text may be biased. 

Are LLMs suitable for evaluating poetry-related tasks? \citet{wang2024best} applied GPT-4 to evaluate the poem's translations using the same rules as human evaluation, but the evaluation results of GPT-4 were far from those of humans, and there were many cases where full marks or close to full marks appeared, proving that GPT-4 is not suitable for the evaluation of poetry translation. 

% This question has not been explored in previous work related to large models and poetry.

% ChatGPT can interpret and analyze poems \cite{chen3using}.

% Therefore, to study the capabilities of LLMs to evaluate poetry-related tasks, 

% Understanding of poetry is the prerequisite for poetry-related works \cite{tate1940understanding,pierce2003defining,mcgrath2018understating}. 

% Therefore, to study the capabilities of LLMs to evaluate the poetry understanding task, our paper explores and enhances the ability of LLMs to evaluate the poetry understanding task. 

Understanding of poetry serves as an essential foundation for all research involving poetry \cite{pierce2003defining, tate1940understanding,mcgrath2018understating}. This paper aims to study and enhance LLMs' ability as evaluators to evaluate the poetry understanding task. Specifically, we first explored LLMs' performance of understanding poetry from multiple dimensions. To that end, we collected 40 high-quality unpublished modern Chinese poems by 5 professional modern Chinese poets, of which 8 poems are written by the same poet. Human evaluation results show that GPT-4 can understand poetry with an accuracy rate of over 76\%. Then, we explored the evaluation capabilities of several mainstream LLMs on ``understanding of poetry'' from multiple dimensions, but the results were not satisfactory because the scores from LLMs were far from those of humans. These initial results evaluated by multiple LLMs are consistent with the results of \citet{wang2024best} evaluating poems' translation with GPT-4, which once again proves that current LLMs are not yet viable as evaluation tools for poetry-related tasks.

% \footnote{In our work, understanding of poetry refers to the interpretation and example analysis of poetry generated by the model from different dimensions.}

% which once again proves that current LLMs are not suitable for evaluating poetry-related tasks. 

% The evaluation results from professional poets show that GPT-4 can correctly understand more than 76\% of the content of poetry. 

% the current nine mainstream LLMs

% we propose a novel evaluation method based on LLMs

To enhance the evaluation ability of LLMs, we propose a novel method called Poller (Poetry LLM Evaluator). Poetry is a special literary genre, and the author of poetry (i.e., poet) usually writes with assumed characters. To accurately understand poetry, readers should first determine the specific role assumed by the author of the poem before reading it \cite{John1938, culler2017theory}. Therefore, our method requires LLMs to play the role of the authors of poems and then evaluate the understanding of poems in the same way as human (real author) evaluation. Specifically, we provide multiple LLMs with detailed information about these authors through prompts, including the basic information of the authors, the authors' own poetry experience and views, and the comments of others (professional poetry critics) on the authors and their published poems. Subsequently, these models evaluate the interpretations of poems from eight dimensions under different prompts and provide reasons for the score. Experimental results demonstrate that our Poller method effectively reduces the evaluation error between LLMs and humans, achieving a 57.37\% reduction in average error for GPT-4-Turbo and 46.95\% for DeepSeek-R1 compared to baseline methods. Especially for Specific dimension evaluation, Poller-based LLMs achieve a 94.55\% reduction in error for rhetorical techniques and an 89.53\% reduction in error for defamiliarization. These performances are unattainable by conventional LLM evaluation methods. Experimental results of multiple LLMs from multiple dimensions validate the efficacy of our method. 

The main contributions of our work include:

%We are the first to investigate large language models (LLMs) as evaluation tools for the foundation of poetry-related tasks (the understanding of poetry).

\begin{itemize}
    \item We are the first to investigate large language models (LLMs) as evaluation tools for the poetry understanding task.
    \item We propose a novel LLMs-based evaluation method, which can be extended to other poetry-related tasks.
   % \item Our method effectively mitigates GPT-4’s bias in favoring its own generation.
%    \item We collected a set of high-quality, unpublished modern Chinese poems, which will be released upon acceptance of the paper.
    \item Experimental results using multiple LLMs across multiple dimensions validate the efficacy of our method.

    % Our work bridges the gap between automated efficiency and human expertise, establishing a foundation for automated evaluation in poetry-related tasks.
\end{itemize}

% We are the first to study the performance of using LLMs to evaluate the foundation of poetry-related tasks: poetry understanding.

\section{Related Work}
Previous work evaluated poetry tasks in different languages and genres. \citet{walsh2024sonnet} evaluated LLMs on the recognition of poetic forms of English poetry, focusing on sonnets, sestinas, and pantoums. Their experimental results demonstrated that when the form of the poems is fixed, LLMs can accurately recognize these poems regardless of whether they are common or uncommon poems. However, LLMs still have difficulties in recognizing poetic forms that are not fixed, such as those based on topic or visual features. \citet{possi2023carmen} proposed a specific metric to automatically evaluate morphological features such as rhyme and meter in poetry, filling the gap in automatic poetry evaluation. Their experiments involved ballad and sonnet and found that the metric is not limited to form, which can distinguish poetry from prose texts (such as news articles). \citet{agirrezabal2023erato} proposed an automatic evaluation framework that can be used for both automatically generated poems and human-written poems. Their evaluation focuses on four aspects of poetry, including poetic features, novelty features, lexico/semantics, and poetic fluency. They conducted experiments on English and Spanish poems and found that, in addition to evaluation, the framework can also be used for the analysis of syllable and stanza structure of sonnets. Another work by \citet{agirrezabal2024zero} evaluated whether Llama2 can generate poetry with predefined metric constraints in zero-shot scenarios. Specifically, they measured how well the generated poetry matched simple constraints, including the number of lines, stanzas, and syllables. Quantitative analysis showed that in zero-shot scenarios, Llama2 had difficulty generating text that followed specific constraints, especially in relation to the number of syllables.

% \citet{possi2023carmen} proposed a specific metric called CARMEN to automatically evaluate morphological features such as rhyme and meter in poetry,

% Quantitative analysis showed that in zero-shot scenarios, Llama2 had difficulty generating text that followed specific constraints, especially in relation to the number of syllables. But the model could infer the correct number of lines or syllables for typical metres.

Obviously, these automatic evaluation works involve poems with features such as rhyme, line number, rhythm, etc., and none of them involve Chinese poetry. Different from these studies, modern Chinese poetry in our work is characterized by freedom and is not constrained by these characteristics \cite{awan2015new, wang2024best}. Due to the open form, diverse genres, and innovative structure of modern Chinese poetry \cite{wang2006guoview, skerratt2013form}, evaluating modern Chinese poetry is challenging. Therefore, to fill the gap in automatic evaluation of modern Chinese poetry, our work focuses on multiple mainstream LLMs and aims to enhance their evaluation performance.

\section{Task Definition}

% The core objective of this work is to enhance LLMs' capability in evaluating poetry understanding through \textbf{author role-playing}

% It is important to emphasize that our evaluation task focuses on assessing the model's comprehension of poetry, rather than the poetry itself.

The core objective of this work is to enhance LLMs' capability in evaluating poetry understanding by playing the role of the poem's author, which simulates a poet's interpretive perspectives. It is worth emphasizing that our task evaluates the model's understanding of poetry, not the poetry itself. Figure \ref{overall workflow} presents the overall workflow of our work. Our methodology comprises three key phases:
%\begin{enumerate}
\begin{enumerate}[label=\arabic*., itemsep=0pt]
    \item \textbf{Multidimensional Understanding Generation}: Prompting LLMs to generate interpretations $U$ of poem $P$ across eight defined dimensions $k$.
    \item \textbf{Human Evaluation Benchmark}: Establishing reference evaluation $H^{(k)}(P, U)$ through human assessment.
    \item \textbf{LLMs Evaluation}: Evaluating the same interpretations using mainstream LLMs under diverse prompting strategies $\theta$.
\end{enumerate}
Evaluation criteria and rules remain identical between human and LLM evaluators. The evaluation deviation is formalized as:
\begin{equation}
\varepsilon_{\theta}^{(k)}(P, U) = f_{\theta}^{(k)}(P, U \mid \mathcal{I}) - H^{(k)}(P, U) 
\label{eq:dim-error}
\end{equation}
where \( k \in \{1,\dots,8\} \) denotes the evaluation dimension index (covering dimensions like content, language, imagery, etc.). \( \theta \) represents the prompt strategies (\(\mathcal{P}_0\) to \(\mathcal{P}_4\)). \(\mathcal{I} = \{I_i, V_i, C_i\}\) encapsulates the detailed information of the poem's author, including basic information \(I_i\), poetic views \(V_i\), and critics' comments \(C_i\). \(\varepsilon_{\theta}^{(k)}\) signifies the evaluation deviation of strategy \(\theta\) for dimension \(k\). And the optimization objective is to minimize the expected absolute deviation:
\begin{equation}
\min_{\theta} \mathbb{E}\left[ \left| \varepsilon_{\theta}^{(k)} \right| \right]
\label{eq:opt-goal}
\end{equation}
This framework optimizes the evaluation error in each dimension and identifies the optimal evaluation strategy \(\theta^*_k\) for specific dimensions.

% The scoring deviation is formalized as:

% This framework enables optimizing the error in each dimension and identifying the optimal evaluation strategy \(\theta^*_k\) for specific dimensions.

% This framework enables:
% \begin{itemize}
%     \item Optimizing the error in each dimension.
%     \item Identifying the optimal evaluation strategy $\theta^*_k$ for specific dimensions.
% \end{itemize}

% \(\mathcal{I} = \{I_i, V_i, C_i\}\) encapsulates the poem's author information including basic details \(I_i\), poetic views \(V_i\), and critics' comments \(C_i\),

% where:
% \begin{itemize}
%     \item $k \in \{1,\dots,8\}$: Evaluation dimension index (content, language, imagery, etc.)
%     \item $\theta$: Prompt strategies ($\mathcal{P}_0$ to $\mathcal{P}_4$)
%     \item $\mathcal{I}$: Detailed information of poem's author (basic information $I_i$, poetic views $V_i$, critics' comments $C_i$)
%     \item $\varepsilon_{\theta}^{(k)}$: Evaluation deviation of strategy $\theta$ for dimension $k$
% \end{itemize}

\section{Evaluation Capabilities of LLMs}
\label{sec: Evaluation Capabilities of LLMs}

% Understanding poetry is the prerequisite for all poetry-related work \cite{tate1940understanding,pierce2003defining,mcgrath2018understating}. To investigate the initial capabilities of LLMs in evaluating poetry understanding tasks, we first prompted LLMs to generate multi-dimensional interpretations of poems. Then, these interpretations were evaluated by human to establish reference benchmarks. Finally, we prompted multiple mainstream LLMs to assess the same interpretations using identical evaluation criteria and rules as the human evaluators. In our work, ``understanding of poetry'' refers to the interpretation and example analysis of poetry generated by the model.

% Therefore, to study the capabilities of LLMs to evaluate poetry understanding task, we first explore the ability of LLMs to evaluate ``understanding of poetry''.

% \cite{manurung2012using, zhang2014chinese, yan2016, yi2017generating, yi2018automatic, refaee2023okaz}

\begin{table*}
\centering
% \small
  \begin{tabular}{lllllllllll}
    \hline
  Evaluator $\downarrow$ &  & Cont         & Lang   & Imag & Rhet
 & Rhyt & Defa  & Thou & Mode & Avg\\
    \hline
    Human & & 80.33  & 79.05 & 81.18  & 77.83  & 76.15 & 79.40 & 78.80 & 79.88  & 79.08  \\ \hline

%    GPT-3.5 & \GP{0} & 94.25	& 89.63	& 94.63	& 89.63	& 88.13	& 89.75	& 94.25	& 93.25  & 91.69 \\
%    GPT-3.5-Poller             & \GP{1} & 93.25	& 88.25	& 93.63	& 88.88 & 	85.25	& 86.13	& 92.75	& 93.00	& 90.14 \\ \hline
    
%    GPT-4o & \GP{0} & 92.38	& 89.38	& 94.00	& 88.50	& 88.00	& 87.38	& 91.75	& 93.25 & 90.58\\
%    GPT-4o-Poller       & \GP{1} & 93.00	& 88.25	& 93.88	& 88.13	& 88.88	& 86.38	& 90.88	& 93.88	& 90.41 \\ \hline
      
    GPT-4 & \GP{0} & 95.25	& 92.25	& 97.00	& 93.13	& 88.75	& 90.38	& 96.50	& 94.38  & 93.46\\
    GPT-4-Poller       & \GP{1} & 90.88	& 89.00	& 93.88	& 89.38	& 86.38	& 83.00	& 92.63	& 91.63	& 89.60 \\ \hline
    
%    o1-preview & \GP{0} & 94.70	& 92.03	& 96.33	& 91.40	& 90.35	& 93.90	& 95.23	& 96.75 & 93.87\\
%    o1-preview-Poller           & \GP{1} & 90.63	& 90.13	& 93.50	& 89.38	& 88.13	& 91.00	& 92.13	& 94.75	& 91.21 \\ \hline

    DeepSeek & \GP{0} & 91.45	& 89.93	& 92.58	& 89.20	& 86.85	& 96.38	& 92.75	& 94.35  & 91.69\\
    DeepSeek-Poller            & \GP{1} & 89.00	& 88.93	& 90.50	& 86.80	& 82.90	& 87.83	& 90.15	& 92.65	& 88.60 \\ \hline

%    Qwen-Max & \GP{0} & 95.00	& 89.13	& 94.13	& 93.00	& 90.00	& 86.13	& 95.00	& 95.00  & 92.17\\
%    Qwen-Max-Poller        & \GP{1} & 91.88	& 87.50	& 93.13	& 89.13	& 83.25	& 88.50	& 93.00	& 93.38	& 89.97 \\ \hline
    
%    Qwen-Plus & \GP{0} & 94.50	& 89.50	& 94.50	& 90.00	& 85.13	& 91.25	& 90.25	& 94.88  & 91.25 \\
%    Qwen-Plus-Poller         & \GP{1} & 91.38	& 88.63	& 95.25	& 91.13	& 86.75	& 88.63	& 94.00	& 94.63	& 91.30 \\ \hline
    
    Qwen-Turbo & \GP{0} & 90.00	& 88.25	& 90.00	& 90.00	& 80.00	& 83.00	& 90.00	& 90.00  & 87.66\\
    Qwen-Turbo-Poller           & \GP{1} & 97.25	& 90.88	& 93.75	& 91.38	& 90.88	& 89.00	& 94.38	& 93.88	& 92.66  \\ \hline
    
    Claude3.5 & \GP{0} & 91.45	& 88.55	& 91.55	& 84.15	& 88.45	& 89.25	& 92.83	& 94.50 & 90.09\\
    Claude3.5-Poller & \GP{1} & 91.90	& 88.00	& 93.73	& 84.95	& 89.35	& 91.88	& 92.68	& 95.18	& 90.96  \\ \hline
    
    \hline
  \end{tabular}
  \caption{\label{Evaluation results of P1 and P2}
    Evaluation results of humans and different models using \GP{0} and \GP{1}. The models in the first column are GPT-4-Turbo, DeepSeek-R1, Qwen2.5-Turbo, and Claude3.5-sonnet. The models that apply our method are named with the suffix Poller. Avg represents the average score of 8 dimensions. The bolded scores are relatively close to human scores.
  }
\end{table*}

\subsection{LLM Generate Understanding of Poetry}
\label{LLM Generate Understanding of Poetry}

\paragraph{Data} Referring to the scale of poetry datasets used in previous studies \cite{wang2024best, virvou2023chatgpt, hutson2023poetry, antar2023effectiveness}, we collected 40 high-quality \textbf{unpublished} modern Chinese poems by 5 professional modern Chinese poets, of which 8 poems are written by the same poet. These poems total 110 stanzas and 600 lines. It is worth mentioning that the poets participating in this study are all senior experts in the field of modern Chinese poetry. They have extensive experience in poetry writing and are familiar with poetry theory.

\paragraph{Dimensions} 
Based on the recommendations of eight professional modern Chinese poets, and referring to previous paper on modern poetry \cite{ChangYinabrieflytalks2012}, we summarized eight key dimensions \cite{yuhaizhang2001, wangguangming1998, XiYunshuTwoCharacteristics, WangShuting2006, WangMi2019, ChengBo2005, wang2024best} for understanding modern Chinese poetry: (1) \textbf{Cont}ent: Summarize what the poem is about. (2) \textbf{Lang}uage: What are the characteristics of the language of this poem? Is it innovative? (3) \textbf{Imag}ery: What imageries are used in this poem, and what are their functions? (4) \textbf{Rhet}orical Techniques: What rhetorical techniques are used in this poem? What is the function? (5) \textbf{Rhyt}hm: How is the rhythm of this poem? Such as pauses, lines, and stanzas. (6) \textbf{Defa}miliarization: Does this poem contain any expression of defamiliarization? If there is any unfamiliar expression, please point it out; if not, you can skip it. (7) \textbf{Thou}ght \& Emotion: What thought and emotion does this poem express? (8) \textbf{Mode}rnity: Is the poem modern or modern-minded? 

These dimensions are clearly delineated to ensure clarity and precision in the evaluation process. Each dimension has its own unique focus:
\begin{itemize}
\item Cont: Focuses on the subject matter or theme of the poem—essentially, what the poem describes.
\item Lang: Refers to the linguistic features of the poem, such as word choice, syntax, and diction.
\item Imag: Examines the use of visual and sensory images in the poem to evoke specific feelings or ideas.
\item Rhet: Focuses on the rhetorical strategies employed, such as metaphor, personification, and other literary techniques.
\item Rhyt: Assesses the internal rhythm of the poem, often expressed in external forms like pauses, line breaks, and stanza structure.
\item Defa: Highlights a common technique in modern poetry that challenges conventional perceptions, emphasizing innovation and originality.
\item Thou: Captures the ideas and emotions conveyed by the poem, focusing on its depth and resonance.
\item Mode: Reflects the contemporary consciousness or modern sensibility embedded in the poem, which distinguishes it as a product of modern literature.
\end{itemize}

% Detailed definitions of all dimensions are shown in the Appendix \ref{subsec:Detailed Definition}. 

Subsequently, GPT-4 (gpt-4-turbo) \cite{achiam2023gpt} generates an understanding of all poems from eight dimensions. 
The prompt and experimental details are shown in Appendix \ref{subsec:Prompts Designed for Evaluating}. 

In this work, we use Chinese prompts directly in the source language without translating them into English. This is because translating Chinese poetry into English for use as prompts introduces several challenges. For instance, this process often results in the loss of key poetic elements such as poeticity, style, cultural nuances, and rhythm \cite{chakrabarty2021don, song2023towards, wang2024best}. These elements are crucial for accurately evaluating a model’s understanding of poetry. To ensure fairness and preserve the integrity of our analysis, we avoided translating Chinese poems into English. Instead, we used the original Chinese poetry directly as input prompts for ChatGPT. This approach ensures that the evaluation reflects the full richness and complexity of the original poems' text without the distortions introduced by translation.

% Subsequently, GPT-4 (gpt-4-turbo) generates the understanding of all poems from eight dimensions, using the prompt shown in Appendix \ref{subsec:Prompts Designed for Evaluating}. Understanding is not a creative task, but understanding of the same content can be described in different ways. So the temperature of the model is set to 0.5 and the top\_p is set to 1.0.

% \begin{table}[t]
% \centering
% \small
% \begingroup
% \begin{tabular}{cccc}
% \hline
% {Poems} & {Stanzas} & {Lines} & {Words} \\
% \hline
% 40 & 110 & 600 & 8K \\
% \hline
% \end{tabular}
% \endgroup
% \caption{\label{poems' data}
% The statistics of poems' data.
% }
% \end{table}

\subsection{Human Evaluation}
\label{Human Evaluation}
Human evaluation solves the disadvantage that automatic evaluation does not consider the intrinsic characteristics of poetry. Reading poetry is inherently subjective, as interpretations often vary significantly among readers \cite{Brodsky1999WitnessandPleasure, JiangYongjun2012,ZhangTaozhou2022Problems}. However, the author's original intent provides a unique benchmark for evaluation. To accurately confirm the model's ability to understand poetry, we invited the five authors of poems to evaluate the experimental results from all dimensions.

% Human evaluation, with the advantage of professional knowledge, solves the disadvantage that automatic evaluation does not consider the intrinsic characteristics of poetry. 

% The detailed definitions of each dimension are shown in Appendix \ref{subsec:Detailed Definition}.

\paragraph{Evaluation Rules} Each dimension ranges from 0 to 100. A score of 0 means that authors think the model's understanding of a certain dimension is completely wrong, and 100 indicates that authors think the model's understanding of a certain dimension is completely correct. The more correct the model's understanding of a certain dimension is, the higher the score authors will give. 

%Higher scores represent a greater overlap between GPT-4's understanding and the original poet's intent.

Evaluation results are reported in the first row of Table \ref{Evaluation results of P1 and P2}. Obviously, the model’s understanding of the poems scored over 76 points in eight dimensions, indicating that the model is able to grasp most of the meaning of the poem.

\begin{table}[t]
% \begin{table}[h]
% \begin{table}[H]
\centering
 \small
% \begin{adjustbox}{width=\}
% \renewcommand{\arraystretch}{0.9}
\begin{tabular}{p{7.2cm}}
\hline
From now on, you are a professional Chinese modern poet and need to play the role of a judge. Your details are as follows:  \\
% \vspace{0.001cm}

\textbf{Basic Information:} \{\{$I_i$\}\}   \\
% Poeticity: Choose the most poetic sentence in this poem.

\textbf{Your Own Experience and Views on Poetry:} \{\{$V_i$\}\}   \\

% If you give 50 points, please provide the sentence you think is the most poetic. 
\textbf{Others' Comments (Professional Poetry Critics) on You and Your Published Poems:} \{\{$C_i$\}\}   \\
% \vspace{0.01cm}
The above are your details. Now, you need to play the role of a professional poet with the above characteristics. Your task is to play the role of a judge. Specifically, you need to score the following ``readers' understanding of the poem'' from the eight dimensions of content, language, imagery, rhetorical techniques, rhythm, defamiliarization, thought and emotion, and modernity, and give detailed reasons for your score. Please note that you only need to score and give reasons for the ``Readers' Understanding of this poem, not the poem itself.  \\
% \vspace{0.01cm}
\textbf{Meaning of Dimension:} Same as human evaluation  \\
% \vspace{0.01cm}
% Meaning of Each Scoring Dimension
% Poeticity: Choose the most poetic sentence in this poem.
\textbf{Scoring Rules:} Each dimension ranges from 0 to 100. A score of 0 means that you think the reader's understanding of a certain dimension is completely wrong, and 100 indicates that you think the reader's understanding of a certain dimension is completely correct; the more correct the reader's understanding of a certain dimension is, the higher the score you will give.  \\
% \vspace{0.01cm}
\textbf{Poem:} \{\{$P_i$\}\}   \\
% \vspace{0.01cm}
\textbf{Readers' Understanding of this Poem:} \{\{$U_i$\}\}  \\
% \vspace{0.01cm}
% \MYFONTSIZE{Poem: \{\{$P_i$\}\} (Directly used the original Chinese poem) } \\
% \MYFONTSIZE{Interpretation of this poem: \{\{$I_i$\}\}}  \\
\textbf{Please remember to state your own role before giving your formal answer.}  \\
\hline
\end{tabular}
\normalsize % 恢复默认字号（避免影响后续内容）
\renewcommand{\arraystretch}{1.0}
% \end{adjustbox}
\caption{\label{The prompt 3 we designed}The Prompt 3 (\GP{3}) we designed for evaluating models' understanding of poetry.
}
\end{table}

% Evaluation results of humans (original poets) and different models using \GP{0} and \GP{1}.

% The specific versions of the models in the first column are detailed in Section \ref{Experiment}.

%The scores are averages of the 40 poems. 

% Human evaluation results and analysis 【？】

\subsection{LLMs Initial Evaluation}
\label{LLMs Initial Evaluation}

% To explore the evaluation ability of LLMs, LLMs were asked to score the understanding of poems generated by GPT-4-Turbo-0125. The content being evaluated and the evaluation rules used are exactly the same as those for human evaluation.

% The evaluation rules used by LLMs are exactly the same as those of human evaluation.

To explore the evaluation capabilities of LLMs, we prompted LLMs to evaluate the interpretations of poems generated by GPT-4-Turbo in Section \ref{LLM Generate Understanding of Poetry}. Specifically, we experimented with several mainstream LLMs, including: GPT-4-Turbo, DeepSeek-R1, Qwen2.5-Turbo, Claude3.5-sonnet \cite{achiam2023gpt, guo2025deepseek, yang2024qwen2, Claude}. The text being evaluated and the evaluation rules used are exactly the same as those for human evaluation. 

% Specifically, we experimented with several mainstream LLMs, including: GPT-3.5-Turbo, GPT-4o, GPT-4-Turbo, o1-preview, DeepSeek-R1, Qwen2.5-Max, Qwen2.5-Plus, Qwen2.5-Turbo, Claude3.5-sonnet \cite{achiam2023gpt, guo2025deepseek, yang2024qwen2, Claude}.

% Specifically, we experimented with nine mainstream LLMs, including: Claude3.5-sonnet-2024-10-22, GPT-3.5-Turbo-0125, GPT-4o-2024-11-20, GPT-4-Turbo-2024-04-09, o1-preview, Qwen2.5-Max, Qwen2.5-Plus, Qwen2.5-Turbo, DeepSeek-R1 \cite{achiam2023gpt, yang2024qwen2, guo2025deepseek}. 

% The evaluation needs to be objective and fair, so the temperature of the evaluation model is set to 0.1. 

LLMs can generate interpretable evaluations and feedback \cite{li2024llms}, and providing explanations can effectively improve the accuracy of LLMs' evaluation \cite{chiang2023closer}. So in addition to scoring, these models were also tasked to provide reasons for the score so that we can fully grasp the models' evaluation performance. Decomposing complex evaluation into multiple dimensions is more conducive to the LLMs' evaluation \cite{song2024finesure}, so the models were required to evaluate ``understanding of poems'' from multiple specific dimensions similar to human evaluation. The prompts are shown in Table \ref{The prompt 0 we designed}, denoted as \GP{0}. \GP{0} is used as the baseline.

% In our prompts, we use Chinese poems directly in the source language without translating them into English. The reasons are shown in Appendix \ref{Chinese Prompts}.

% \item \textbf{\GP{0}}: \underline{Same as the rules for human evaluation. Details are shown in Table \ref{The prompt 1 we designed}.}

% \GP{0} in Table \ref{Evaluation results of P1 and P2} present the evaluation results of LLMs on the ``understanding of poems'' from different dimensions.

% \GP{0} in Table \ref{Evaluation results of P1 and P2} present the evaluation results of LLMs from different dimensions. The results are unexpected, as we are disappointed to see that LLMs are not reliable for all dimensions. Specifically, Qwen2.5-Turbo performed best relative to the other eight models, as its average score (87.66) is closest to the human average score (79.08). In terms of specific dimensions, Qwen2.5-Turbo's score for defamiliarization (83.00) is closest to the human score (79.40), and its score for rhythm is only 3.85 points different from the human score. However, for other dimensions, the difference between Qwen2.5-Turbo's score and the human score is more than 8.82 points.

\GP{0} in Table \ref{Evaluation results of P1 and P2} present the evaluation results of LLMs from different dimensions. The results are unexpected, as we are disappointed to see that LLMs are not reliable for all dimensions. Specifically, Qwen2.5-Turbo performed best relative to the other eight models, as its average score (87.66) is closest to the human average score (79.08). In terms of specific dimensions, Qwen2.5-Turbo's score for defamiliarization (83.00) is closest to the human score (79.40). However, for other dimensions, the difference between Qwen2.5-Turbo's score and the human score is more than 8.82 points.

% The average score of the other models except Qwen2.5-Turbo for poetry understanding is at least 11.01 points different from the human score (Claude3.5-sonnet), and the score for specific dimensions is at least 6.32 points different from the human score (Claude3.5-sonnet). The model with the worst evaluation performance is o1-preview, whose average score for poetry understanding (93.84) is 14.76 points different from the human score (79.08). Among them, o1-preview has the worst evaluation performance of all models for rhythm and modernization. For GPT-4-Turbo, although we used a different model version from generating understanding, which performed second worst, as the gap between its score and human scores was second only to o1-preview. We speculate that this is because GPT-4 is able to recognize and favor its own generated text, which is consistent with the conclusion of previous works \cite{liu2024llms, bitton2023visit, koo2023benchmarking, panickssery2024llm}.

The average score of the other models except Qwen2.5-Turbo for poetry understanding is at least 11.01 points different from the human score (Claude3.5-sonnet), and the score for specific dimensions is at least 6.32 points different from the human score (Claude3.5-sonnet). The model with the worst evaluation performance is GPT-4-Turbo, whose average score (93.46) for poetry understanding is 14.38 points different from the human score (79.08). Among them, GPT-4-Turbo has the worst evaluation performance of all models for content, language, imagery, rhetorical techniques, rhythm, and 
thought and emotion. Furthermore, we noted that the evaluation scores given by GPT-4-Turbo are significantly higher than those given by humans. We speculate that this is because GPT-4 is able to recognize and favor its own generated text, which is consistent with the conclusion of previous works \cite{liu2024llms, bitton2023visit, koo2023benchmarking, panickssery2024llm}.

% 除了Qwen2.5-Turbo之外，其他模型在诗歌理解方面的平均得分与人类得分（Claude3.5-sonnet）至少相差11.01分，在特定维度上的得分与人类得分（Claude3.5-sonnet）至少相差6.32分。评估表现最差的模型是o1-preview，其诗歌理解的平均得分（93.84）与人类得分（79.08）相差14.76分。其中，o1-preview在韵律和现代化方面的表现也是所有模型中最差的。对于GPT-4-Turbo，尽管我们使用了与生成理解不同的模型版本（其表现排名倒数第二），但其得分与人类得分之间的差距仅次于o1-preview。我们推测这是因为 GPT-4 能够识别并偏爱自己生成的文本，这与先前的工作的结论一致 \cite{liu2024llms, bitton2023visit, koo2023benchmarking, panickssery2024llm}。

% \paragraph{LLMs tend to give higher scores than humans} 
We observed that for all dimensions, all LLMs gave higher evaluation scores than humans in the initial evaluation (\GP{0}). For the evaluation of modernity, humans gave a score of 79.88, but all models gave scores (\GP{0}) higher than or equal to 90; among them, GPT-4-Turbo, DeepSeek-R1, and Claude3.5-sonnet all gave evaluation scores (\GP{0}) higher than 93.25. Similarly, for the evaluation of content, humans gave a score of 80.33, but all models gave scores (\GP{0}) higher than or equal to 90. We speculate that this may be because LLMs focus on surface linguistic features (such as the use of novel words) while ignoring the cultural context inside the text. This is consistent with recent findings that LLMs' processing of metaphor may rely on superficial lexical associations rather than reliable semantic understanding \cite{ye2025unveiling}.

Overall, the evaluation results from all LLMs (all dimensions) are not close to those of human evaluation, proving that current LLMs are not suitable for the evaluation of poetry understanding task.

\section{Poetry LLM Evaluator}

% Specifically, LLMs first play the role of the original poet and then evaluate the ``understanding of poetry''. 

% En-LLM-Eval (\textbf{En}hance \textbf{LLMs} \textbf{Eval}uation).

To enhance the evaluation ability of LLMs, we propose a novel LLM-based method termed Poller (\textbf{Po}etry \textbf{LL}M \textbf{E}valuato\textbf{r}). Poller requires LLMs to play the role of five authors of poems and then evaluate the understanding of poems in the same way as human evaluation. Specifically, we give LLMs detailed information about these authors through prompts, including: (1) Authors' basic information, such as occupation, growth background, life background, personal experience, and so on; (2) Authors' own experiences and views on poetry.; and (3) Others' comments (professional poetry critics) on authors and their published poems. Subsequently, the model evaluates the understanding of the poems from eight dimensions under different prompts and gives reasons. 

% Figure \ref{En-LLM-Eval framework} presents the framework of our proposed Poller method.

\paragraph{LLM-evaluators Take the Author's Perspective} When a poet writes a poem, he will enter a specific state that is different from his daily life. The poet in this state is a kind of assumed character \cite{culler2017theory, John1938}. Therefore, the expectation for LLMs to adopt the author’s perspective is grounded in the relationship between the cultural context, the author’s intent, and the deeper understanding of the poem:

\begin{itemize}
\item In any era, an author’s perspective is influenced by the cultural, historical, and social context in which they live. This perspective significantly impacts the style, form, and the thoughts and emotions the author seeks to convey through their poetry \cite{DiZangpoetic}. For example, the imagery and rhetorical techniques chosen by the author often reflect their personal experiences and worldview.

% For example, the themes, imagery, and rhetorical techniques chosen by the author often reflect their worldview and personal experiences.

\item By adopting the author’s perspective, evaluators can better understand the intentions behind the poem and the cultural or emotional subtext embedded within it. This approach allows evaluators to go beyond a superficial reading and grasp the deeper meaning and artistic choices of the author.
\item Evaluators who adopt the author’s perspective are better equipped to assess a poem’s intended impact and artistic merit. This is particularly important in the context of modern Chinese poetry, where cultural nuances and historical influences play a crucial role in shaping the poet’s creative process and output.
\end{itemize}

For LLMs, different prompts may lead to different evaluation results \cite{xu2023llm}.
Therefore, to explore LLMs' ability of evaluating the ``understanding of poetry'', we designed several prompts, denoted as $\mathcal{P}_{i\in\{1...4\}}$, with the following focuses:

% For LLMs, different prompt templates may lead to different evaluation results \cite{xu2023llm}.

\begin{itemize}
\item \textbf{\GP{1}}: \textit{Based on \GP{0}, model plays the role of the poems' author and serves as a judge. Compared with \GP{0}, \GP{1} contains: (1) basic information about the original poet, (2) the original poet's experience and views on poetry, and (3) comments from others (professional poetry critics) on the original poet's published poems.} 
\item \textbf{\GP{2}}: \textit{Based on \GP{1}, it emphasizes that the poem is written by the model (playing the poet), and the understanding of the poem comes from other readers.} 
\item \textbf{\GP{3}}: \textit{Based on \GP{1}, the model is required to declare its role before the formal scoring.} 
\item \textbf{\GP{4}}: \textit{It combines the characteristics of \GP{2} and \GP{3}, emphasizing that the poem is written by the model (playing the poet), while the understanding of the poem comes from other readers. At the same time, the model is required to declare its role before the formal scoring.} 
\end{itemize}
% Details are shown in Table \ref{The prompt 4 we designed}.
Table \ref{The prompt 3 we designed} presents the Prompt 3 (\GP{3}) we designed for evaluating models' understanding of poetry. Other prompts and complete details are provided in the Appendix \ref{subsec:Prompts Designed for Evaluating}.

% Higher scores represent a greater overlap between GPT-4’s understanding and the original poet’s intent.

\begin{table*}[t]
  \centering
%  \small
  \begin{tabular}{llcccccccccc}
    \hline
    Evaluator $\downarrow$ &  & Cont         & Lang   & Imag & Rhet
 & Rhyt & Defa  & Thou & Mode & Avg\\
    \hline

% GPT-4系列
GPT-4 & $\mathcal{P}_0$ & 14.92 & 13.20 & 15.82 & \textit{15.30} & 12.60 & \textit{10.98} & 17.70 & 14.50 & 14.38 \\
GPT-4-Poller & $\mathcal{P}_1$ & 10.55 & 9.95 & 12.70 & 11.55 & 10.23 & 3.60 & 13.83 & 11.75 & 10.52 \\
GPT-4-Poller & $\mathcal{P}_2$ & 14.57 & 11.83 & 15.82 & 16.67 & 13.85 & 8.93 & 16.25 & 15.22 & 14.14 \\
GPT-4-Poller & $\mathcal{P}_3$ & 9.67 & 5.95 & 7.82 & \textit{2.30} & \underline{3.23} & \underline{\textbf{\textit{1.15}}} & 8.33 & 10.62 & \underline{6.13} \\
GPT-4-Poller & $\mathcal{P}_4$ & 10.30 & 7.20 & 10.82 & 7.80 & 8.98 & 2.73 & 9.45 & 12.50 & 8.72 \\ 

\midrule

% DeepSeek系列
DeepSeek & $\mathcal{P}_0$ & 11.12 & 10.88 & 11.40 & 11.37 & 10.70 & 16.98 & 13.95 & 14.47 & 12.61 \\
DeepSeek-Poller & $\mathcal{P}_1$ & 8.67 & 9.88 & 9.32 & 8.97 & 6.75 & 8.43 & 11.35 & 12.77 & 9.52 \\
DeepSeek-Poller & $\mathcal{P}_2$ & 10.10 & 10.13 & 6.50 & 5.65 & 8.98 & 8.35 & 9.30 & 13.80 & 9.10 \\
DeepSeek-Poller & $\mathcal{P}_3$ & 9.00 & 5.88 & \underline{5.82} & \underline{\textbf{0.62}} & 6.83 & 5.40 & \underline{7.48} & 12.47 & 6.69 \\
DeepSeek-Poller & $\mathcal{P}_4$ & 12.27 & 9.75 & 9.45 & 7.45 & 11.43 & 9.90 & 10.88 & 15.42 & 10.81 \\

\midrule

% Qwen-Turbo系列
Qwen-Turbo & $\mathcal{P}_0$ & 9.67 & 9.20 & 8.82 & 12.17 & 3.85 & 3.60 & 11.20 & 10.12 & 8.58 \\
Qwen-Turbo-Poller & $\mathcal{P}_1$ & 16.92 & 11.83 & 12.57 & 13.55 & 14.73 & 9.60 & 15.58 & 14.00 & 13.59 \\
Qwen-Turbo-Poller & $\mathcal{P}_2$ & 8.42 & 7.58 & 8.57 & 9.80 & 9.85 & 6.23 & 10.70 & 10.75 & 8.99 \\
Qwen-Turbo-Poller & $\mathcal{P}_3$ & 9.55 & 3.33 & 8.20 & 5.42 & 5.48 & 1.60 & 11.08 & 8.12 & 6.60 \\
Qwen-Turbo-Poller & $\mathcal{P}_4$ & \underline{6.55} & 5.45 & 8.70 & 8.55 & 6.23 & 3.73 & 11.08 & 10.12 & 7.55 \\

\midrule

% Claude3.5系列
Claude3.5 & $\mathcal{P}_0$ & 11.12 & 9.50 & 10.37 & 6.32 & 12.30 & 9.85 & 14.03 & 14.62 & 11.01 \\
Claude3.5-Poller & $\mathcal{P}_1$ & 11.57 & 8.95 & 12.55 & 7.12 & 13.20 & 12.48 & 13.88 & 15.30 & 11.88 \\
Claude3.5-Poller & $\mathcal{P}_2$ & 7.45 & 9.53 & 6.85 & 6.05 & 14.03 & 6.10 & 8.28 & 13.72 & 9.00 \\
 Claude3.5-Poller & $\mathcal{P}_3$ & 6.90 & \underline{2.15} & 6.90 & 2.25 & 8.50 & 4.23 & 8.88 & 11.40 & 6.40 \\
 Claude3.5-Poller & $\mathcal{P}_4$ & 8.72 & 8.25 & 10.35 & 4.85 & 11.88 & 5.10 & 11.53 & 13.45 & 9.27 \\ 

\hline
  \end{tabular}
  \caption{\label{Evaluation errors}
    Evaluation errors (absolute deviation from human scores) of different models and prompting strategies. Values represent absolute errors: $|\text{Model Score} - \text{Human Score}|$. Italics indicate the optimal evaluation performance under this dimension, that is, the absolute error between the LLM evaluation result and the human evaluation result is the smallest.
  }
\end{table*}

% The models that apply our method are named with the suffix Poller. Avg represents the average score of 8 dimensions. The bold scores represent the scores closest to humans in different dimensions.

\section{Experiment}
\label{Experiment}

\paragraph{Authors' Information} To effectively stimulate the ability of LLMs to act as authors, we collected and screened the detailed information of all authors. Specifically, we collected a total of 37 poetry-related articles from the 5 original poets from interviews, speeches, poets' seminars, poets' published poetry collections, and published journals. Among them, 13 articles are the original poets' own experiences and views on poetry, and 24 articles are professional poetry critics' comments on the poems (not overlapping with the poems in our experiment) of the poets. We finally obtained 57K words of poetry-related information about the 5 original poets, and summarized the personal information of each poet to control the number of words between 1K and 2.5K words. % The statistics of authors' information are shown in Table \ref{information data}. 

To determine the potential of LLMs to play the role of poems' authors (\GP{1}), we conduct experiments using the same LLMs as in Section \ref{LLMs Initial Evaluation}.
The results of the eight dimensions are reported in Table \ref{Evaluation results of P1 and P2}. Table \ref{Evaluation results of P1 and P2} shows that the model with the greatest potential for playing the role of the author to evaluate ``understanding of poetry'' is GPT-4-Turbo, because its average score after playing the role of the author (89.60) is 3.86 points lower than the average score when it does not play the role of the author (93.46), which is the largest reduction among all models. In addition, when playing the role of the poem' author (\GP{1}), DeepSeek-R1's score (88.60) is closest to human.

% In addition, when playing the role of the poem' author (\GP{1}), DeepSeek-R1's score (88.60) is closest to humans, because its difference (9.52) with human scores is relatively the smallest.

Based on the initial evaluation performance (Table \ref{Evaluation results of P1 and P2}) of using \GP{0} and their potential to act as poems' authors using \GP{1}, we selected the four most representative models to continue the experiments using \GP{2}, \GP{3}, and \GP{4}. The four models in the subsequent experiments are: Claude3.5-sonnet, GPT-4-Turbo, Qwen2.5-Turbo, and DeepSeek-R1. 

Table \ref{Evaluation errors} presents the evaluation errors (absolute deviation from human scores) of different models and prompting strategies. Figure \ref{Example picture} shows an example of our Poller method. Complete examples and more evaluation results are provided in Appendix \ref{Experimental Examples} and \ref{Prompt Each Dimension Separately}, respectively.

% \paragraph{Results} 

% Table \ref{Evaluation results of P3 to P5} presents the evaluation results of LLMs on the understanding of poems by applying our Poller method (\GP{1} to \GP{4}) on four models and without applying our method (\GP{0}). 

% Other metrics used to measure the relationship between LLMs and human scores are shown in Appendix \ref{Other metrics}. For example, we calculated the Mean Absolute Error (MAE) as a supplement to further quantify the difference between LLMs evaluations and human evaluations. The MAE results in Table \ref{Mean Average Error} in Appendix \ref{Other metrics} confirm the effectiveness of our proposed method, Poller, in reducing the error between LLM and human evaluations across multiple dimensions. This verifies that Poller can improve the evaluation performance of LLM. 

% This once again verifies that Poller can improve the evaluation performance of LLM. 

% Based on the initial evaluation performance (Table \ref{Evaluation results of P1 and P2}) of all models using \GP{0} and their potential to act as poems' authors using \GP{1}

% The contents of the first line are the abbreviations of Claude3.5-sonnet, GPT-4-Turbo, Qwen2.5-Turbo, and DeepSeek-R1.

% 【Examples of evaluation results of rhetorical techniques by different models】

\section{Analysis}

\paragraph{Our Poller method effectively narrows the gap between LLMs and human evaluation.} 
\begin{itemize}
    \item Overall: As shown in Table \ref{Evaluation errors}, after using our method, the overall evaluation error of LLMs is reduced compared to the initial evaluation error without using our method (\GP{0}). For example, compared with the baseline prompt strategy (\GP{0}), Poller's optimal strategy (\GP{3}) reduces the average evaluation error of GPT-4-Turbo from 14.38 to 6.13 (↓57.37\%), DeepSeek-R1 from 12.61 to 6.69 (↓46.95\%), Claude3.5-sonnet from 11.01 to 6.40 (↓41.87\%), and Qwen2.5-Turbo from 8.58 to 6.60 (↓23.08\%). 
   \item Single dimension: For \textbf{rhetorical techniques}, the evaluation error of DeepSeek-R1 using \GP{3} is only 0.62, close to the human level, and reduced by 94.55\% compared with the baseline \GP{0} (11.37); the evaluation error of GPT-4-Turbo using \GP{3} (2.30) is reduced by 84.97\% compared with \GP{0} (15.30), and the evaluation error of Sonnet using \GP{3} (2.25) is reduced by 64.40\% compared with \GP{0} (6.32). Similarly, for \textbf{defamiliarization}, the evaluation error of GPT-4-Turbo using \GP{3} (1.15) is reduced by 89.53\% compared with \GP{0} (10.98), while the evaluation error of DeepSeek-R1 using \GP{3} (5.40) is reduced by 68.20\% compared with \GP{0} (16.98).
\end{itemize}

The reduction in the error between LLMs evaluation and human evaluation proves that our Poller method effectively enhances the evaluation ability of LLMs. Poller injects detailed information about the poet through role-playing, guiding LLMs to simulate the author's intention, thereby alleviating the bias caused by the lack of cultural context in the LLMs' traditional evaluation.

\paragraph{The effectiveness of the various prompting strategies we designed.} The optimal evaluation performance of the large model on different dimensions appears on the different prompting strategies we designed. As shown in Table \ref{Evaluation errors}, the optimal evaluation performance for \textbf{content} appears when Qwen2.5-Turbo uses \GP{4} in our Poller method, and its absolute error with the human evaluation result is 6.55. The optimal evaluation performance for \textbf{language} appears when Claude3.5-sonnet uses \GP{3} in our Poller method, and its absolute error with the human evaluation result is 2.15. Interestingly, we found that except for Qwen2.5-Turbo, which uses \GP{4} to obtain the best performance in content evaluation, the evaluation performance of all other models on all dimensions is best when using \GP{3}.

These results demonstrate the effectiveness of our design of multiple prompting strategies. Specifically, both the \GP{3} and \GP{4} strategies in the Poller method require the model to declare its role before starting formal scoring, which shows that the behavior of the model confirming its role strengthens the identity of LLMs, effectively improving its ability to play the role of the poem author, thereby effectively improving its evaluation performance.

% \paragraph{LLMs tend to give higher scores than humans.} During the experiment, we observed that for all dimensions, all LLMs gave higher evaluation scores than humans in the initial evaluation (\GP{0}) without using our method. As shown in Table \ref{Evaluation results of P1 and P2}, for the evaluation of modernity, the score given by humans is 79.88, but the scores given by all models (\GP{0}) are higher than or equal to 90 points; among them, the evaluation scores (\GP{0}) given by GPT-3.5-Turbo, GPT-4o, GPT-4-Turbo, o1-preview, DeepSeek-R1, Qwen2.5-Max, Qwen2.5-Plus, and Claude3.5-sonnet are all higher than 93.25 points. Similarly, for the evaluation of content, the score given by humans is 80.33, but the scores given by all models (\GP{0}) are higher than or equal to 90 points.

% We speculate that this may be because LLMs focus on surface language features (such as the use of novel words) and ignore the cultural context inside the text.

\paragraph{Can LLMs be used as evaluation tools?} The experimental results (Table \ref{Evaluation errors}) show that the model using our method is more reliable in the evaluation task than the model not using our method (\GP{0}). Therefore, LLMs using our method can be used to assist in evaluation. Specifically, DeepSeek-R1 is suitable for evaluating rhetorical techniques when using \GP{3}, GPT-4-Turbo is suitable for evaluating defamiliarization and rhythm when using \GP{3}, and Claude3.5-sonnet is suitable for evaluating language when using \GP{3}.

\begin{table}[t]
  \centering
  \begin{tabular}{lcc}
    \hline
    &  All Dimensions &  Single Dimension \\
    \hline
    Rhet & 78.45	& 79.00 \\
    Defa & 78.25	& 86.50 \\
    \hline
  \end{tabular}
  \caption{\label{Result of Prompt Each Dimension Separately}
    The evaluation scores from the LLMs when GPT-4 understands all dimensions and single dimension. The numbers in the second column refer to the evaluator's score for a specific dimension when GPT-4 generates poetry understanding from all dimensions simultaneously. The numbers in the third column refer to the score of generating poetry understanding from only a single dimension.  
  }
\end{table}

\section{Prompt Each Dimension Separately}
\label{Prompt Each Dimension Separately}

To explore whether it is better to prompt each dimension separately, we have added additional experiments on two separate dimensions, Rhetorical Techniques and Defamiliarization. Taking Rhetorical Techniques as an example, our experimental process is divided into two steps:

\begin{itemize}

    \item  Prompt GPT-4 to generate understanding of 20 poems only from the Rhetorical Techniques dimension.
    \item Use LLMs (DeepSeek-Poller) whose evaluation performance on the Rhetorical Techniques dimension is close to that of human evaluation to score the understanding generated in the first step through \GP{3}.

\end{itemize}

Based on the results and analysis of Table \ref{Evaluation results of P3 to P5} in our paper, we use DeepSeek-Poller (\GP{3}) as the evaluator of Rhetorical Techniques and GPT-4-Poller (\GP{3}) as the evaluator of Defamiliarization. The evaluation scores from the LLMs are shown in Table \ref{Result of Prompt Each Dimension Separately}. 

We found that when the single dimension was prompted separately, the model obtained higher scores in both Rhetorical Techniques and Defamiliarization. This shows that prompting a specific dimension alone can enhance the model's understanding of poetry. But in reality, evaluating the quality of a poem requires considering multiple dimensions rather than a single one. Therefore, this work focuses on evaluating poetry from multiple dimensions.

\section{Conclusion}

Existing automatic evaluation methods are not suitable for poetry domain, while human evaluation is expensive. In this paper, we propose a novel evaluation method based on LLMs, which can effectively evaluate poetry understanding task with performance closer to human evaluation results than traditional LLM evaluation. This method requires LLMs to play the role of the poem's author and evaluate the understanding of poems from multiple dimensions. Experimental results demonstrate that our method effectively reduces the evaluation error between LLMs and humans, especially in the two dimensions of evaluating rhetorical techniques and defamiliarization. Experimental results using multiple LLMs across multiple dimensions validate the efficacy of our method. Our work bridges the gap between automated efficiency and human expertise, establishing a foundation for automated evaluation in poetry-related tasks.

% Existing automatic evaluation methods are not suitable for poetry domain, while human evaluation is expensive. In this paper, we propose a novel evaluation method based on LLMs, which can effectively evaluate poetry understanding task with performance close to human evaluation results. Specifically, our method requires LLMs to play the role of authors of poems with detailed information and evaluate the understanding of poems from multiple dimensions. Experimental results show that the evaluation performance of LLMs using our method is better than the initial evaluation performance without our method. In addition, the evaluation results of LLMs using our method are significantly closer to human evaluation results, especially in the two dimensions of evaluating rhetorical techniques and defamiliarization. The experimental results of multiple LLMs from multiple dimensions prove the effectiveness of our proposed method. Our work lays the foundation for using LLMs to evaluate other poetry-related tasks in the future.

\section*{Limitations}

% Our method requires LLMs to act as the authors of poems who is given detailed information (basic information, the author's experience and views on the poem, and comments from professional poetry critics on the author's published poems). The authors have authorized the use of this information in our research. Our method can be generalized to other related tasks based on LLMs, but the author information collected in our work is not applicable to non-poetry tasks.

Our work focuses solely on modern Chinese poetry, a genre characterized by free form. While the proposed Poller method effectively reduces the evaluation error of LLMs, its applicability to other poetic forms remains to be explored. The role-playing mechanism relies on poet-specific background information, which may be difficult to obtain in other traditional texts. Nevertheless, we believe that the core idea of adopting the author's perspective is transferable, as adapting Poller to new domains would require primarily re-sourcing poet information rather than redesigning the evaluation framework. This limitation does not undermine the validity of our findings for modern Chinese poetry.

% Referring to the scale of poetry datasets used in previous studies, we collected 40 high-quality unpublished modern Chinese poems written by several professional modern Chinese poets. These poems total 110 stanzas, 600 lines, and 8K words (excluding titles). 

% Compared to existing works on large language models (LLMs) and poetry, our dataset is not small. For example:

% \begin{itemize}
%    \item \citet{antar2023effectiveness} used 1 poem (10 lines) in their evaluation.
%    \item \citet{hutson2023poetry} focused on 4 modern poems.
%    \item \citet{wang2024best} used 10 poems in their main experiments.
%    \item \citet{virvou2023chatgpt} evaluated 30 poems.
%\end{itemize}

%Thus, our dataset is among the largest in studies related to LLMs and poetry. By including a larger and more diverse set of poems, we aim to enhance LLMs' evaluation performance on poetry understanding task.

%However, compared to non-poetry related tasks, the amount of our data is relatively small, although our data is of high quality as it comes from professional poets.

\section*{Ethics Statement}
 
The human poems used in this study are derived from the authors of these poems, and all authors have authorized our work to use the poem data.

\section*{Acknowledgements}

This work was supported in part by the Science and Technology Development Fund of Macau SAR (Grant Nos. FDCT/0007/2024/AKP, EF2024-00185-FST), the UM and UMDF (Grant Nos. MYRG-GRG2024-00165-FST-UMDF, MYRG-GRG2025-00236-FST), the Tencent AI Lab Rhino-Bird Research Program (Grant No. EF2023-00151-FST), the Stanley Ho Medical Development Foundation (Grant No. SHMDF-AI/2026/001), and the National Natural Science Foundation of China (Grant No. 62266013).

\bibliography{main}

\appendix

\section{Appendix}
\label{sec:appendix}

\subsection{Prompts}
\label{subsec:Prompts Designed for Evaluating}

\paragraph{LLM Generate Understanding} Table \ref{The prompt designed for LLMs to understand modern poetry} presents the prompt we designed for LLMs to understand modern Chinese poetry. Understanding is not a creative task, but understanding the same content can be described in different ways. So the temperature of the model is set to 0.5, and the top\_p is set to 1.0.

\paragraph{LLM Evaluation} The prompts (\GP{0} to \GP{4}) we designed for evaluating the ``understanding of poetry'' are shown in Table \ref{The prompt 0 we designed} to \ref{The prompt 4 we designed}, respectively. The evaluation needs to be objective and fair, so the temperature of the evaluation model is set to 0.1.

\begin{table}[t]
% \begin{table}[h]
% \begin{table}[H]
\centering
\small
% \begin{adjustbox}{width=\}
\begin{tabular}{p{7.2cm}}
\hline
Please understand the following modern poem titled $T_i$ from the five dimensions of content, expression methods, thought and emotion, modernity, and the most poetic sentence, and finally give your interpretation of the poem. In addition to content, please provide examples and specific explanations for each dimension. Please note that this is a complete modern poem. The specific definitions of each dimension are as follows:  \\
%\vspace{0.001cm}
Content: Summarize what the poem is about.  \\
Expression Methods: Understand how this poem is written from five aspects: language, imagery, rhetorical techniques, rhythm, and defamiliarization. The five specific aspects are as follows: 1) What are the characteristics of the language of this poem? Is it innovative? 2) What imageries are used in this poem, and what are their functions? 3) What rhetorical techniques are used in this poem? What is the function? 4) How is the rhythm of this poem? Such as pauses, lines, and stanzas. 5) Does this poem contain any expression of defamiliarization? If there is any unfamiliar expression, please point it out; if not, you can skip it.  \\
Thought \& Emotion: What thoughts and emotions does this poem express?  \\
Modernity: Is the poem modern or modern-minded?  \\
% \vspace{0.01cm}
Poeticity: Choose the most poetic sentence in this poem. \\
Poem: \{\{$P_i$\}\} (Directly used the original Chinese poem)  \\
Interpretation of this poem: \{\{$I_i$\}\}  \\

\hline

\end{tabular}
% \end{adjustbox}
\caption{\label{The prompt designed for LLMs to understand modern poetry}The prompt we designed for LLMs to understand modern Chinese poetry.
}
\end{table}

\begin{table}[t]
% \begin{table}[h]
% \begin{table}[H]
%\centering
% \begin{adjustbox}{width=\}
%\renewcommand{\arraystretch}{0.9}
\small
\begin{tabular}{p{7.2cm}} 
\hline
Please score the following ``readers' understanding of the poem'' from the eight dimensions of content, language, imagery, rhetorical techniques, rhythm, defamiliarization, thought and emotion, and modernity, and give detailed reasons for your score. Please note that you only need to score and give reasons for the ``Readers' Understanding of this poem, not the poem itself.  \\
% \vspace{0.001cm}
\textbf{Meaning of Dimension:} Same as human evaluation  \\
% \vspace{0.001cm}
% Poeticity: Choose the most poetic sentence in this poem.
\textbf{Scoring Rules:} Each dimension ranges from 0 to 100. A score of 0 means that you think the reader's understanding of a certain dimension is completely wrong, and 100 indicates that you think the reader's understanding of a certain dimension is completely correct; the more correct the reader's understanding of a certain dimension is, the higher the score you will give.  \\
% \vspace{0.001cm}
% If you give 50 points, please provide the sentence you think is the most poetic. 
\textbf{Poem:} \{\{$P_i$\}\}   \\
% \vspace{0.001cm}
\textbf{Readers' Understanding of this Poem:} \{\{$U_i$\}\}  \\
% \vspace{0.01cm}
% \MYFONTSIZE{Poem: \{\{$P_i$\}\} (Directly used the original Chinese poem) } \\
% \MYFONTSIZE{Interpretation of this poem: \{\{$I_i$\}\}}  \\
\hline
\end{tabular}
% \end{adjustbox}
\caption{\label{The prompt 0 we designed}The Prompt 0 (\GP{0}) we designed for evaluating models' understanding of poetry.}
\end{table}

\begin{table}[t]
% \begin{table}[h]
% \begin{table}[H]
\small
\begin{tabular}{p{7.2cm}} 
\hline
From now on, you are a professional Chinese modern poet and need to play the role of a judge. Your details are as follows:  \\
%\vspace{0.001cm}
\textbf{Basic Information:} \{\{$I_i$\}\}   \\
% Poeticity: Choose the most poetic sentence in this poem.
\textbf{Your Own Experience and Views on Poetry:} \{\{$V_i$\}\}   \\
% If you give 50 points, please provide the sentence you think is the most poetic. 
\textbf{Others' Comments (Professional Poetry Critics) on You and Your Published Poems:} \{\{$C_i$\}\}   \\
The above are your details. Now, you need to play the role of a professional poet with the above characteristics. Your task is to play the role of a judge. Specifically, you need to score the following ``readers' understanding of the poem'' from the eight dimensions of content, language, imagery, rhetorical techniques, rhythm, defamiliarization, thought and emotion, the most poetic sentence, and modernity, and give detailed reasons for your score. Please note that you only need to score and give reasons for the ``Readers' Understanding of this poem, not the poem itself.  \\

\textbf{Meaning of Dimension:} Same as Prompt 0  \\
\textbf{Scoring Rules:} Same as Prompt 0  \\

\textbf{Poem:} \{\{$P_i$\}\}   \\
\textbf{Readers' Understanding of this Poem:} \{\{$U_i$\}\}  \\
% \vspace{0.01cm}
% \MYFONTSIZE{Poem: \{\{$P_i$\}\} (Directly used the original Chinese poem) } \\
% \MYFONTSIZE{Interpretation of this poem: \{\{$I_i$\}\}}  \\

\hline

\end{tabular}
% \end{adjustbox}
\caption{\label{The prompt 1 we designed}The Prompt 1 (\GP{1}) we designed for evaluating models' understanding of poetry.
}
\end{table}

\begin{table}[t]
% \begin{table}[h]
% \begin{table}[H]
\small
\begin{tabular}{p{7.2cm}} 
\hline
From now on, you are a professional Chinese modern poet and need to play the role of a judge. Your details are as follows:  \\
%\vspace{0.001cm}
\textbf{Basic Information:} \{\{$I_i$\}\}   \\
% Poeticity: Choose the most poetic sentence in this poem.
\textbf{Your Own Experience and Views on Poetry:} \{\{$V_i$\}\}   \\
% If you give 50 points, please provide the sentence you think is the most poetic. 
\textbf{Others' Comments (Professional Poetry Critics) on You and Your Published Poems:} \{\{$C_i$\}\}   \\
The above are your details. \textbf{You recently wrote the following modern poem, which has not been published yet, and some readers have understood and interpreted this modern poem. Please note that the following poem is written by you, and the understanding of your poem comes from other readers.} Now, you need to play the role of a professional poet with the above characteristics. Your task is to play the role of a judge. Specifically, you need to score the following ``readers' understanding of the poem'' from the eight dimensions of content, language, imagery, rhetorical techniques, rhythm, defamiliarization, thought and emotion, the most poetic sentence, and modernity, and give detailed reasons for your score. Please note that you only need to score and give reasons for the ``Readers' Understanding of this poem, not the poem itself.  \\

\textbf{Meaning of Dimension:} Same as Prompt 1  \\
\textbf{Scoring Rules:} Same as Prompt 1  \\

\textbf{Poem:} \{\{$P_i$\}\}   \\
\textbf{Readers' Understanding of this Poem:} \{\{$U_i$\}\}  \\
% \vspace{0.01cm}
% \MYFONTSIZE{Poem: \{\{$P_i$\}\} (Directly used the original Chinese poem) } \\
% \MYFONTSIZE{Interpretation of this poem: \{\{$I_i$\}\}}  \\

\hline

\end{tabular}
% \end{adjustbox}
\caption{\label{The prompt 2 we designed}The Prompt 2 (\GP{2}) we designed for evaluating models' understanding of poetry.
}
\end{table}

\begin{table}[t]
% \begin{table}[h]
% \begin{table}[H]
\small
\begin{tabular}{p{7.2cm}} 
\hline
From now on, you are a professional Chinese modern poet and need to play the role of a judge. Your details are as follows:  \\
%\vspace{0.001cm}
\textbf{Basic Information:} \{\{$I_i$\}\}   \\
% Poeticity: Choose the most poetic sentence in this poem.
\textbf{Your Own Experience and Views on Poetry:} \{\{$V_i$\}\}   \\
% If you give 50 points, please provide the sentence you think is the most poetic. 
\textbf{Others' Comments (Professional Poetry Critics) on You and Your Published Poems:} \{\{$C_i$\}\}   \\
The above are your details. \textbf{You recently wrote the following modern poem, which has not been published yet, and some readers have understood and interpreted this modern poem. Please note that the following poem is written by you, and the understanding of your poem comes from other readers.} Now, you need to play the role of a professional poet with the above characteristics. Your task is to play the role of a judge. Specifically, you need to score the following ``readers' understanding of the poem'' from the eight dimensions of content, language, imagery, rhetorical techniques, rhythm, defamiliarization, thought and emotion, the most poetic sentence, and modernity, and give detailed reasons for your score. Please note that you only need to score and give reasons for the ``Readers' Understanding of this poem, not the poem itself.  \\

\textbf{Meaning of Dimension:} Same as Prompt 1  \\
\textbf{Scoring Rules:} Same as Prompt 1  \\

\textbf{Poem:} \{\{$P_i$\}\}   \\
\textbf{Readers' Understanding of this Poem:} \{\{$U_i$\}\}  \\
% \vspace{0.01cm}
% \MYFONTSIZE{Poem: \{\{$P_i$\}\} (Directly used the original Chinese poem) } \\
% \MYFONTSIZE{Interpretation of this poem: \{\{$I_i$\}\}}  \\
\textbf{Please remember to state your own role before giving your formal answer.}  \\

\hline

\end{tabular}
% \end{adjustbox}
\caption{\label{The prompt 4 we designed}The Prompt 4 (\GP{4}) we designed for evaluating models' understanding of poetry.
}
\end{table}

\subsection{Evaluation Results of Poller}
\label{Evaluation Results of Poller}

Table \ref{Evaluation results of P3 to P5} presents the evaluation results of applying our Poller method (\GP{1} to \GP{4}) on four models and without applying our method (\GP{0}).

\begin{table*}
  \centering
  \begin{tabular}{lllllllllll}
    \hline
    Evaluator $\downarrow$ &  & Cont         & Lang   & Imag & Rhet
 & Rhyt & Defa  & Thou & Mode & Avg\\
    \hline   
    Human & & 80.33  & 79.05 & 81.18  & 77.83  & 76.15 & 79.40 & 78.80 & 79.88  & 79.08  \\ \hline

     GPT-4 & \GP{0} & 95.25	& 92.25	& 97.00	& 93.13	& 88.75	& 90.38	& 96.50	& 94.38  & 93.46\\
     GPT-4-Poller      & \GP{1} & 90.88	& 89.00	& 93.88	& 89.38	& 86.38	& 83.00	& 92.63	& 91.63	& 89.60 \\  
     GPT-4-Poller       & \GP{2} & 94.90	& 90.88	& 97.00	& 94.50	& 90.00	& 88.33	& 95.05	& 95.10	& 93.22 \\
     GPT-4-Poller           & \GP{3} & 90.00	& 85.00	& 89.00	& 80.13	& 79.38	& \textbf{78.25}	& 87.13	& 90.50	& 84.92 \\
     GPT-4-Poller           & \GP{4} & 90.63	& 86.25	& 92.00	& 85.63	& 85.13	& 82.13	& 88.25	& 92.38	& 87.80 \\ \hline

    DeepSeek & \GP{0} & 91.45	& 89.93	& 92.58	& 89.20	& 86.85	& 96.38	& 92.75	& 94.35  & 91.69\\
    DeepSeek-Poller            & \GP{1} & 89.00	& 88.93	& 90.50	& 86.80	& 82.90	& 87.83	& 90.15	& 92.65	& 88.60 \\
    DeepSeek-Poller             & \GP{2} & 90.43	& 89.18	& 87.68	& 83.48	& 85.13	& 87.75	& 88.10	& 93.68	& 88.18 \\
    DeepSeek-Poller             & \GP{3} & 89.33	& 84.93	& 87.00	& \textbf{78.45}	& 82.98	& 84.80	& 86.28	& 92.35	& 85.77 \\
    DeepSeek-Poller             & \GP{4} & 92.60	& 88.80	& 90.63	& 85.28	& 87.58	& 89.30	& 89.68	& 95.30	& 89.90 \\ \hline

    Qwen-Turbo & \GP{0} & 90.00	& 88.25	& 90.00	& 90.00	& 80.00	& 83.00	& 90.00	& 90.00  & 87.66\\
     Qwen-Turbo-Poller           & \GP{1} & 97.25	& 90.88	& 93.75	& 91.38	& 90.88	& 89.00	& 94.38	& 93.88	& 92.66  \\
     Qwen-Turbo-Poller          & \GP{2} & 88.75	  &  86.63  &  89.75 &   87.63	& 86.00	& 85.63	& 89.50	& 90.63	& 88.07 \\
    Qwen-Turbo-Poller           & \GP{3} & 89.88	& 82.38	& 89.38 & 	83.25	& 81.63	& 81.00	& 89.88	& 88.00	& 85.68 \\
     Qwen-Turbo-Poller          & \GP{4} & 86.88	& 84.50	& 89.88	& 86.38	& 82.38	& 83.13	& 89.88	& 90.00	& 86.63 \\ \hline

    Claude3.5 & \GP{0} & 91.45	& 88.55	& 91.55	& 84.15	& 88.45	& 89.25	& 92.83	& 94.50 & 90.09\\
     Claude3.5-Poller   & \GP{1} & 91.90	& 88.00	& 93.73	& 84.95	& 89.35	& 91.88	& 92.68	& 95.18	& 90.96  \\ 
    Claude3.5-Poller & \GP{2} & 87.78	 &  88.58	&   88.03	&  83.88	&   90.18	&   85.50	&   87.08	&   93.60	&   88.08\\
     Claude3.5-Poller         & \GP{3} & 87.23	& 81.20	& 88.08	& 80.08	& 84.65	& 83.63	& 87.68	& 91.28	& 85.48\\
    Claude3.5-Poller          & \GP{4} & 89.05	& 87.30	& 91.53	& 82.68	& 88.03	& 84.50	& 90.33	& 93.33	& 88.34 \\ \hline

  \end{tabular}
  \caption{\label{Evaluation results of P3 to P5}
    Evaluation results of applying our Poller method (\GP{1} to \GP{4}) on four models and without applying our method (\GP{0}). 
  }
\end{table*}

\subsection{Experimental Examples}
\label{Experimental Examples}

% An example of our proposed Poller (Poetry LLM Evaluator) method.

\subsubsection{The Example Without Using Our Method} 

An example of Prompt 0 (\GP{0}) without using our method is as follows:

\textbf{Poem:} 
%(Figure \ref{Chinese Poem Example figure} presents the original Chinese version of this poem)

\begin{center}
\begin{CJK}{UTF8}{gkai}醉酒一事\end{CJK}

(The Matter of Being Drunk)\\\

\begin{CJK}{UTF8}{gkai}在酒的国度里，鱼鳌可以\end{CJK}

(In the kingdom of wine, turtles might)

\begin{CJK}{UTF8}{gkai}离开海洋，鸟雀不必站在树上\end{CJK}

(Depart the sea, and birds need not perch in trees)

\begin{CJK}{UTF8}{gkai}男人醉酒后，很放心地\end{CJK}

(A man, inebriated, with carefree trust)

\begin{CJK}{UTF8}{gkai}把自己摆到了街边。人间之事\end{CJK}

(Lay himself along the street’s edge, where worldly matters)

\begin{CJK}{UTF8}{gkai}全都碎在杯里，一饮而尽后\end{CJK}

(All shattered in the cup, drained in one gulp)

\begin{CJK}{UTF8}{gkai}在肠胃里激荡，像巨浪冲刷着堤防\end{CJK}

(Surging in the stomach, like giant waves washing over the embankment)\\\

\begin{CJK}{UTF8}{gkai}车辆在街灯的重影里轰鸣、拉扯\end{CJK}

(Vehicles roar and tug in the doubled shadows of streetlights)

\begin{CJK}{UTF8}{gkai}犹如那些轻易给出的诺言，在黑暗中\end{CJK}

(Like promises too lightly made, in dark)

\begin{CJK}{UTF8}{gkai}徒劳地撒网，没人会在意一个醉汉\end{CJK}

(Casting nets in vain. No one cares about a drunk)

\begin{CJK}{UTF8}{gkai}倒置的身体和语言，他只是马路上的一块礁石\end{CJK}

(Inverted body and language, he is just a rock on the road)

\begin{CJK}{UTF8}{gkai}是礁石上一条搁浅的鱼，鱼腹里的一根刺\end{CJK}

(It's a stranded fish on the reef, a thorn in its belly)\\\

\begin{CJK}{UTF8}{gkai}想到这里，他真的像一根鱼刺一样\end{CJK}

(Thinking of this, he really looks like a fishbone)

\begin{CJK}{UTF8}{gkai}卡住了道路的喉咙\end{CJK}

(Stuck in the throat of the road)
\end{center}

\textbf{GPT-4's understanding of this poem:}

(1) Content: This poem describes the state of a drunkard and his feelings in this state. The poem depicts how alcohol can make people detach from reality. The drunkard seems to be in a "kingdom of wine" that is not bound by the real world, but in the end, this behavior of escaping reality turns him into a reef in the city, a stranded fish, and even a fishbone in the throat of the road. 

(2) Language: The poem uses direct and concrete language, innovatively comparing the drunkard to a reef and a fishbone. This direct metaphor is a common language feature of modern poetry. 

(3) Imagery: The poem uses images such as "fish claws", "birds", "huge waves", and "vehicles", comparing the state of drunkenness to the freedom of animals and the power of natural disasters, which enhances the expressiveness of the poem. 

(4) Rhetorical Techniques: The poem uses rhetorical techniques such as metaphor (such as comparing the drunkard to a reef and a fishbone) and exaggeration (such as "all broken in the cup") to enhance the expressiveness and appeal of the language. 

(5) Rhythm: The poem uses short lines and pauses to control the rhythm, simulating the drunk's intermittent stream of consciousness and his wandering in the city.

(6) Defamiliarization: The poem achieves the effect of defamiliarization by comparing people to crops (reefs, fish bones), allowing readers to see the image and state of the drunk from a new perspective.

(7) Thought \& Emotion: This poem expresses the poet's deep understanding of the state of drunkenness and his sympathy for the drunk's isolation and helplessness in society. Through the image of the drunk, the poet reflects the possible loss and loneliness of people in modern society.

(8) Modernity: This poem has modernity. It takes the experience of an ordinary person in modern society as its starting point, uses the language and expression of modern poetry, and explores the isolation of individuals in society and the complexity of their inner world. Through the depiction of the drunk, the poet reflects the inner struggle of modern people and their desire for freedom.

\textbf{Evaluation results from GPT-4 as the evaluator (\GP{0}) for “GPT-4-Turbo’s understanding of this poem”:}

(1) Content: 100 points

The reader accurately understood that the poem explores the isolation and loss of individuals in society by describing the state of drunkenness.

(2) Language: 90 points

The reader correctly pointed out that the poem uses direct and figurative language, but did not further explore how language enhances the poem's emotional expression.

(3) Imagery: 100 points

The reader accurately identified images such as "fish claws" and "birds" and understood how these images enhance the poem's expressiveness.

(4) Rhetorical Techniques: 100 points

The reader correctly identified and explained the use of rhetorical techniques such as metaphor and hyperbole, and how they enhance the poem's appeal.

(5) Rhythm: 90 points

The reader's description of the poem's rhythm is appropriate, noting the control of sentence length and pauses, but the reader's analysis of how the rhythm fits the theme is not in-depth enough.

(6) Defamiliarization: 100 points

The reader correctly pointed out that the poem achieves the effect of defamiliarization by comparing people to objects (reefs, fish bones).

(7) Thought and Emotion: 100 points

The reader deeply understood the poet's profound understanding of the state of drunkenness and his sympathy for the drunk's isolation and helplessness in society.

(8) Modernity: 100 points

The reader accurately pointed out the modern characteristics of the poem, including its reflection of the inner struggle of modern people and their desire for freedom.

\subsubsection{The Example Using Our Method}

% An example of using our Poller method (\GP{3}) is as follows:

\textbf{Evaluation results from GPT-4 as the evaluator using our method Poller (\GP{3}) for “GPT-4-Turbo’s understanding of this poem”:}

\textbf{Role:} Professional poet, university professor

\textbf{Scores and Reasons}

(1) Content: 90 points

The reader's understanding of the content of the poem is relatively accurate, capturing the state and feelings of the drunkard and the core image of the poem, "the kingdom of wine." The reader also understands the drunkard's isolation in the city.

(2) Language: 85 points

The reader correctly points out the poem's use of direct and figurative language and recognizes the innovativeness of the metaphor. Although the reader mentions that this is a common feature of modern poetry, the reader does not delve into the uniqueness and innovation of the language.

(3) Imagery: 80 points

The reader identifies the main images in the poem and understands the role of these images. However, the reader does not fully explore how these images specifically enhance the expressiveness of the poem.

(4) Rhetorical Techniques: 90 points

The reader accurately identifies the use of metaphor and hyperbole and understands the role these figures of speech play in enhancing the expressiveness of the language.

(5) Rhythm: 75 points

The reader noticed the rhythm control of short lines and pauses, but did not deeply analyze how these rhythm changes specifically simulate the flow of consciousness and state of the drunkard.

(6) Defamiliarization: 85 points

The reader correctly identified the expression of defamiliarization, which provides a new perspective by comparing people to objects. However, the reader did not fully explore how this defamiliarization affects the reader's emotions and understanding.

(7) Thought \& Emotion: 90 points

The reader has a relatively deep understanding of the poem's thoughts and emotions, identifying the poet's sympathy for the drunkard and his reflection on the state of isolation in modern society.

(8) Modernity: 95 points

The reader has a very good understanding of the poem's modernity, identifying the poem's modernity in language and expression, as well as its exploration of the state of individual isolation.

\textbf{Summary}

The reader's understanding of this poem is generally accurate and in-depth, especially in the analysis of thoughts and emotions and modernity. If the reader could have conducted a more in-depth discussion on the uniqueness of the language and the specific role of rhythm, the score would have been higher.

\end{document}